\documentclass[letterpaper, 10 pt, conference]{ieeeconf}  

\IEEEoverridecommandlockouts                              

\usepackage{graphicx}
\usepackage{algorithm}
\usepackage{algorithmic}
\usepackage{xcolor}
\usepackage{booktabs}
\usepackage{multirow}
\usepackage{balance}
\usepackage{amsmath}
\usepackage{bm}
\usepackage{subcaption}
\usepackage[font=small,labelfont=small]{caption}
\usepackage{amssymb}
\begin{document}
\title{\LARGE \bf
Towards Safe Imitation Learning \\via Potential Field-Guided Flow Matching
}

\author{Haoran Ding$^{1}$, Anqing Duan$^{1}$, Zezhou Sun$^{1}$, Leonel Rozo$^{2}$, No\'emie Jaquier$^{3}$,\\ Dezhen Song$^{1}$, Yoshihiko Nakamura$^{1}$
\thanks{
This work was supported by Multi-Institutional Faculty Interdisciplinary Research Project (MFIRP) between IIT Delhi and MBZUAI.
}
\thanks{
$^{1}$Haoran Ding, Anqing Duan, Zezhou Sun, Dezhen Song and Yoshihiko Nakamura are with the Department of Robotics, Mohamed bin Zayed University of Artificial Intelligence (MBZUAI), Abu Dhabi, UAE. {\tt\small \{haoran.ding, anqing.duan, zezhou.sun, dezhen.song, yoshihiko.nakamura\}@mbzuai.ac.ae}
}
\thanks{
$^{2}$Leonel Rozo is with Bosch Center for Artificial Intelligence, Renningen, Germany. {\tt\small leonel.rozo@de.bosch.com}
}
\thanks{
$^{3}$No\'emie Jaquier is with the Division of Robotics, Perception, and Learning, KTH Royal Institute of Technology, Stockholm, Sweden. {\tt\small jaquier@kth.se}
}
} 

\maketitle
\thispagestyle{empty}
\pagestyle{empty}

\begin{abstract}
Deep generative models, particularly diffusion and flow matching models, have recently shown remarkable potential in learning complex policies through imitation learning. However, the safety of generated motions remains overlooked, particularly in complex environments with inherent obstacles. In this work, we address this critical gap by proposing Potential Field-Guided Flow Matching Policy (PF2MP), a novel approach that simultaneously learns task policies and extracts obstacle-related information, represented as a potential field, from the same set of successful demonstrations. During inference, PF2MP modulates the flow matching vector field via the learned potential field, enabling safe motion generation. By leveraging these complementary fields, our approach achieves improved safety without compromising task success across diverse environments, such as navigation tasks and robotic manipulation scenarios. We evaluate PF2MP in both simulation and real-world settings, demonstrating its effectiveness in task space and joint space control. Experimental results demonstrate that PF2MP enhances safety, achieving a significant reduction of collisions compared to baseline policies. This work paves the way for safer motion generation in unstructured and obstacle-rich environments.
\end{abstract}

\section{INTRODUCTION}


Imitation learning (IL)~\cite{imitationlearning} has seen transformative advancements with the introduction of generative models~\cite{kitdiffpol, diffusionpolicy, braun2024riemannian}. Among these, diffusion models~\cite{ddpm,ddim} and flow matching models~\cite{flowmatching,rfm} have emerged as powerful tools, enabling scalable and efficient policy learning. By capturing complex, high-dimensional, and multi-modal distributions, these methods have established generative models as the foundation of next-generation IL frameworks~\cite{deepgenrobo}.
\begin{figure}[t]
    \centering
    \includegraphics[width=\linewidth]{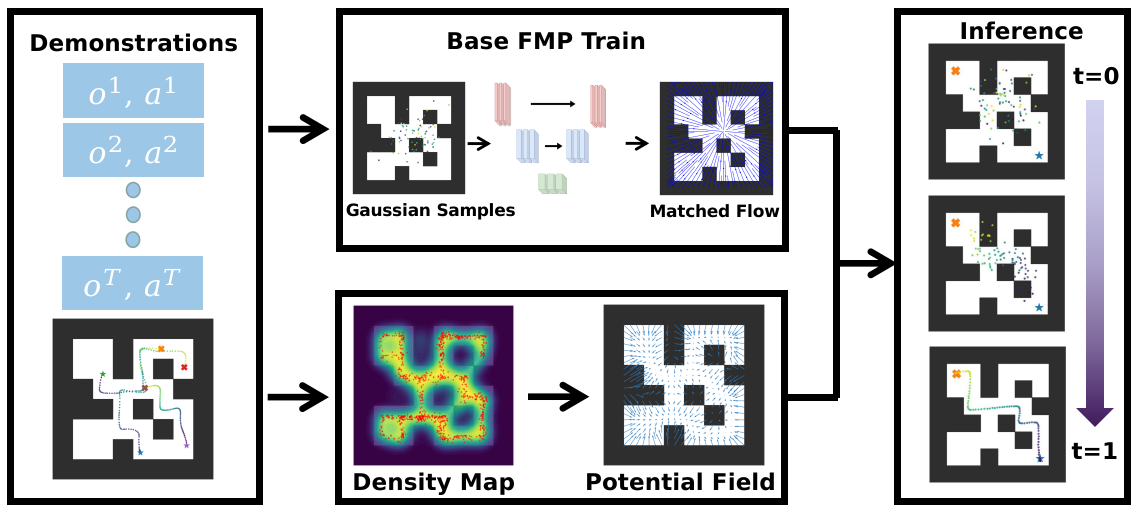}
    \caption{Overview of the PF2MP Framework: The pipeline begins with demonstration data (left). The Base Flow Matching Policy (FMP) is trained (middle top) to model expert behaviors, while a potential field is simultaneously learned (middle bottom) using a density map. During the inference phase (right), the PF2MP framework integrates the learned vector field from FMP and the obstacle-aware potential field to generate safe trajectories.}
    \label{fig:overview of the method}
\end{figure}

Despite these advances, a critical aspect remains insufficiently explored: the safety of the generated policies~\cite{diffplansurvey}. Safety is a critical consideration in real-world applications such as autonomous driving~\cite{safead} and robotic manipulation~\cite{brunke2022safe}, where the consequences of unsafe actions can be catastrophic. 
Recent efforts have begun to incorporate safety considerations into generative models, with notable progress in diffusion-based IL~\cite{colddiffusion,edmp}. However, similar advancements in Flow Matching-based IL are mainly absent.
Current Flow Matching-based IL primarily focus on accurately modeling and reproducing expert demonstrations, often overlooking potential hazards or obstacles inherent in the environment. This gap presents a significant challenge for deploying these methods in safety-critical domains.

To address this limitation, we propose a novel approach to enhance the safety of Flow Matching policies by incorporating obstacle-aware potential fields into the policy generation process. The key insight of our method is to leverage demonstration data not only to learn the desired policy but also to design a potential field derived from the density of the demonstrations estimated via Kernel Density Estimation (KDE).
By guiding the flow matching inference process with the estimated potential field, our approach adjusts trajectories in response to environmental hazards encountered during demonstrations, effectively bridging efficient policy generation with safety assurance.

In summary, we enhance Flow Matching-based IL by incorporating an estimated potential field that accounts for the inherent obstacles in the environment, resulting in the \textbf{P}otential \textbf{F}ield-Guided \textbf{F}low \textbf{M}atching \textbf{P}olicy (\textbf{PF2MP}), illustrated in Fig~\ref{fig:overview of the method}. In contrast to existing diffusion model-based IL~\cite{colddiffusion, edmp, motionplandiffusion}, which rely on manually-defined safety objective functions, our approach autonomously derives safety constraints directly from expert demonstrations. Through extensive experiments in both simulated and real-world environments, we demonstrate that our model effectively improves the safety of the learned policy while preserving task performance.

\section{RELATED WORK}
Ensuring safety is a critical challenge in robot policy learning, with most studies focusing on reinforcement learning~\cite{brunke2022safe, 10478841} and imitation learning~\cite{colddiffusion, safegil, 8624934, menda2019ensembledagger, yin2021imitation}. Recently,
generative models~\cite{ddpm, ddim, flowmatching} have made transformative advancements to robot policy learning, particularly in the field of imitation learning. Here we review recent developments, categorizing them into two major approaches: Diffusion-Based Robot Policies and Flow Matching-Based Robot Policies. 

\subsection{Diffusion-Based Robot Policies}
Diffusion models~\cite{yang2023diffusion}, which iteratively transform a Gaussian distribution into a complex target distribution through denoising steps, have achieved impressive success in various application domains, including image generation~\cite{ddpm}, video synthesis~\cite{videodiffusion}, and human motion modeling~\cite{humanmotiondiffusion}, among others. 
In robotics, Diffusion Policy (DP)~\cite{diffusionpolicy}, primarily trained via Denoising Diffusion Probabilistic Model (DDPM), demonstrated improved performance in imitation learning compared to earlier behavioral cloning methods~\cite{lstm-gmm, ibc, bet} with an enhanced ability to capture multimodal action distributions. Consistency Policy (CP)~\cite{consistencypolicy} addresses the expensive inference process of DP by distilling a student policy from a teacher DP, effectively reducing computational overhead. Furthermore, diffusion models have demonstrated strong performance in complex robot policy learning tasks~\cite{m2diffuser}. For example, 3D Diffusion Policy (DP3)~\cite{DP3} combines a compact 3D visual representation with diffusion models, enabling effective policy learning in more intricate environments.

Several works proposed enhanced diffusion-based policies to generate safer robot motions. Motion Planning Diffusion (MPD)~\cite{motionplandiffusion} guides the diffusion process based on the gradients of multiple objective functions, thereby enhancing the smoothness and safety of the generated trajectories. Similarly, Ensemble-of-costs-guided Diffusion for Motion Planning (EDMP)~\cite{edmp} incorporates the gradients of multiple collision cost functions during denoising, significantly enhancing the safety of generated motions. Cold Diffusion with Replay Buffers (CDRP)~\cite{colddiffusion} augments DP by recording the intermediate variables generated during the denoising process on the training dataset and storing them as a safe replay buffer. During testing, CDRP projects the current denoising intermediate variables onto this buffer, effectively anchoring the inference process to observed safe states. This approach enables CDRP to produce robotic trajectories with reduced collision rate. 

\subsection{Flow Matching-Based Robot Policies}
Flow Matching (FM)~\cite{flowmatching, rectifiedflow} is another class of generative models that transform a simple prior distribution (e.g., a Gaussian distribution) into a complex target distribution via a vector field. In contrast to diffusion models, the FM vector field induces straighter paths and its flow corresponds to an ordinary differential equation (ODE). Therefore, FM is easier to train and faster to query during inference when compared to diffusion models.

Capitalizing on these advantages, several works recently proposed to leverage FM in robot imitation learning. Braun \emph{et al.} introduced Riemannian Flow Matching Policy (RFMP)~\cite{braun2024riemannian} to learn sensorimotor policies represented by end-effector pose trajectories and ensuring geometric constraints in the generated policies. Stable Riemannian Flow Matching Policy (SRFMP)~\cite{haoran2024srfmp} enhances the robustness and inference speed of RFMP by stabilizing the convergence of the flow to the target distribution. ActionFlow~\cite{actionflow} employs FM with a $\mathrm{SE}(3)$ flow and an equivariant transformer to learn $\mathrm{SE}(3)$-equivariant policies that effectively handle relative positions between the end-effector and objects. FM was also applied to multi-support manipulation tasks in~\cite{multisupportfmp}, to control a bimanual robot interacting with multiple environmental supports. Additionally, Chisari \emph{et al.}~\cite{3dfmp} proposed to employed 3D point clouds instead of images as observations, which resulted in improved FM policies performances.

While FM-based methods methods offer efficiency and flexibility, none of these works addressed the safety of the generated FM policies. However, ensuring safe policies is critical in real-world robotics applications, especially in environments with inherent obstacles and dynamic constraints. In this paper, we aim to extend the capabilities of FM-based robot policies by learning not only complex policies from expert demonstrations, but also potential fields to avoid unsafe regions and collisions with the obstacles observed in the demonstrations. 

\section{BACKGROUND}
In this section, we provide a short background on Flow Matching and Kernel Density Estimation (KDE), which serve as the foundational frameworks for our method.

\subsection{Flow Matching}
Flow Matching (FM)~\cite{flowmatching} is a deep generative model that transforms a simple base distribution $p_0(\bm{x})$, typically a Gaussian distribution, into an unknown target distribution $q(\bm{x}_1)$ through a learned vector field. The probability path is generated by push-forwarding $p_0$ along the resulting flow, which is governed by an ordinary differential equation (ODE). While the exact form of the target distribution $q(\bm{x}_1)$ is unknown, it is assumed that samples $\bm{x}_1$ drawn from $q(\bm{x}_1)$ are accessible.

Conditional flow matching (CFM)~\cite{flowmatching} constructs a conditional probability path $p_t(\bm{x}_t|\bm{x}_1)$ linking a Gaussian base distribution $p_0(\bm{x})$ and the target distribution $q(\bm{x}_1)$ by conditioning on the available samples $\bm{x}_1$. The probability path is defined as, 
\begin{equation}
p_t(\bm{x}|\bm{x}_1) = \mathcal{N}(\bm{x}|\mu_t(\bm{x}_1), \sigma_t(\bm{x}_1)^2 \bm{I}),
\end{equation}
where $\mu_t(\bm{x}_1)$ and $\sigma_t(\bm{x}_1)$ are time-dependent mean and scalar standard deviation, respectively. The flow $\psi_t$ that pushes the base distribution to the target distribution can be expressed as,
\begin{equation}
\psi_t(\bm{x}|\bm{x}_1) = \sigma_t(\bm{x}_1) \bm{x} + \mu_t(\bm{x}_1),
\end{equation}
and describes the trajectory of the variable $\bm{x}_t$ as it moves from the base distribution to the target distribution. Lipman \emph{et al.}~\cite{flowmatching} introduce a general formulation of the conditional vector field $u_t$ that generates such flow $\psi_t$,
\begin{equation}
    u_t(\bm{x}|\bm{x}_1) = \frac{\sigma '_t(\bm{x}_1)}{\sigma_t(\bm{x}_1)}(\bm{x}-\mu_t(\bm{x}_1)) + \mu '_t(\bm{x}_1),
    \label{eq:CFM-vector-field}
\end{equation}
where $\mu'_t(\bm{x}_1)$ and $\sigma'_t(\bm{x}_1)$  are the derivatives of the mean and standard deviation functions. By integrating the vector field~\eqref{eq:CFM-vector-field} with initial points $\bm{x}_0$ under a predefined time boundary, the result aligns with the target distribution. The training process of CFM can be framed as a regression problem. A neural network is trained to mimic the conditional vector field $u_t(\bm{x}|\bm{x}_1)$ by minimizing a mean squared error loss,
\begin{equation}
    \mathcal{L}_{\text{CFM}}(\bm{\bm{\theta}}) = \mathbb{E}_{t, q(\bm{x}_1), p(\bm{x}_0)}||v_\theta(\bm{x}_t, t) - u_t(\bm{x}|\bm{x}_1)||^2 ,
\end{equation}
where $v_{\bm{\theta}}$ is the learned vector field, $\bm{\theta}$ represents the parameters of the neural network and $t$ is uniformly sampled within the time boundary. 

Liu \emph{et al.}~\cite{rectifiedflow} proposed a more direct approach by simplifying the conditional vector field to
\begin{equation}
    u_t(\bm{x}|\bm{z}) = \bm{x}_1 - \bm{x}_0,
\end{equation}
where $\bm{z}=(\bm{x}_0, \bm{x}_1)$. This simplified vector field is conditioned not only on the sample $\bm{x}_1$ from the target distribution but also on the sample $\bm{x}_0$ from the base distribution. By connecting the two distributions with a straight path, this approach significantly reduces the computational complexity of the vector field calculation and accelerates the inference process~\cite{rectifiedflow}. 
PF2MP leverages~\cite{rectifiedflow} to train a base Flow Matching Policy to mimic the complex expert policy (see Section~\ref{subsec: base fmp}).

\subsection{Kernel Density Estimation}
Kernel Density Estimation (KDE)~\cite{kde} is a non-parametric method for estimating the probability density function (PDF) of a random variable based on a finite sample of observations. Unlike parametric methods that assume that the data follows a specific distribution (e.g., Gaussian or exponential), KDE makes minimal assumptions about the underlying data distribution, making it a versatile and robust technique in statistical analysis and machine learning.

KDE seeks to estimate the density function $f(\bm{x})$ of a random variable $\bm{X}$, given $M$ independent and identically distributed samples ${\bm{x}_1, \bm{x}_2, ..., \bm{x}_M}$. The KDE kernel estimator is expressed as,
\begin{equation}
    \hat{f}(\bm{x}) = \frac{1}{M h} \sum_{i=1}^M K \left( \frac{\bm{x} - \bm{x}_i}{h} \right),
\end{equation}
where $K(\cdot)$ is the kernel function, i.e., a symmetric, non-negative, and integrable function that determines the shape of the distribution estimate, $h$ is a positive scalar representing bandwidth or smoothing parameter that controls the width of the kernel function, and thus the smoothness of the estimated density function~\cite{densityestimation}. KDE is employed to estimate the density of the demonstrations, which is subsequently used to construct the potential field in PF2MP (see Section~\ref{subsec: pf}).

\section{POTENTIAL FIELD-GUIDED FLOW MATCHING POLICY}
In this section, we introduce the \textbf{P}otential \textbf{F}ield-Guided \textbf{F}low \textbf{M}atching \textbf{P}olicy (\textbf{PF2MP}), illustrated in Fig~\ref{fig:overview of the method}. PF2MP is composed of two components trained from the same set of demonstrations: (1) a base Flow Matching Policy (Section~\ref{subsec: base fmp}); and (2) an estimated potential field (Section~\ref{subsec: pf}). By combining these two components during the inference process (Section~\ref{subsec: pf}), PF2MP generates safer policies that incorporate the environmental constraints observed in the demonstrations. 

\subsection{Base Flow Matching Policy}
\label{subsec: base fmp}
\begin{algorithm}[tp]
\linespread{1.15}\selectfont 
\caption{Base Flow Matching Policy Training Process}
\renewcommand{\algorithmicrequire}{\textbf{Input:}}
\renewcommand{\algorithmicensure}{\textbf{Output:}}
\label{algorithm: training}
\begin{algorithmic}[1]
\REQUIRE Initialized parameter $\bm{\theta}$,  prior distribution $p_0$, demonstration $D=\{\bm{o}^\tau, \bm{a}^\tau\}_{\tau=1}^T$, training epochs $N_T$
\ENSURE Parameter $\bm{\theta}$ of the learned vector field
\FOR{$i=1:N_T$}
    \STATE Sample a time step $t$ from uniform distribution $\mathcal{U}[0, 1]$;
    \STATE Sample a noise $\bm{A}_0$ from prior distribution $p_0$;
    \STATE Sample a target action series $\bm{A}_1$ and the paired observation $\bm{o}$ from the demonstrations;
    \STATE Build the target vector field $u_t$~\eqref{equation:fmp vector field};
    \STATE Calculate the loss function~\eqref{equation: fmp loss function};
    \STATE Update the parameters $\bm{\theta}$;
\ENDFOR
\RETURN $\bm{\theta}$
\end{algorithmic}
\end{algorithm}
This section introduces Flow Matching Policy (FMP), as presented in prior works~\cite{braun2024riemannian, actionflow, multisupportfmp, 3dfmp, haoran2024srfmp, zhang2024flowpolicy}, which constitutes the base policy of PF2MP. 

We consider a set of demonstrations $D=\{\bm{o}^\tau, \bm{a}^\tau\}_{\tau=1}^T$, where $\bm{o}^\tau$ represents the environment observation, $\bm{a}^\tau$ denotes the corresponding action taken by the agent, and $T$ is the total length of the trajectory. The objective of imitation learning is to train a policy $\pi_{\bm{\theta}}(\bm{a};\bm{o})$ that mimics the expert policy $\pi_e(\bm{a};\bm{o})$, which generated the demonstrations. 

Flow Matching Policy (FMP) leverages a flow matching model --- specifically based on~\cite{rectifiedflow} in this paper --- to learn an observation-conditioned vector field $u_t(\bm{A}|\bm{z}, \bm{o}^\tau)$, defined as
\begin{equation}
    \label{equation:fmp vector field}
    u_t(\bm{A}|\bm{z}, \bm{o}^\tau) = \bm{A}_1 - \bm{A}_0,
\end{equation}
where $t\in[0,1]$ represents the time step during the generation process, $\bm{A}$ denotes the intermediate variable during the generation process, the variable $\bm{z}=(\bm{A}_0, \bm{A}_1)$ includes a sample $\bm{A}_0$ from a prior distribution (typically a Gaussian distribution) with the same dimensionality as $\bm{A}_1$ and the target action series $\bm{A}_1=\{\bm{a}^\tau, \bm{a}^{\tau + 1},..., \bm{a}^{\tau + T_a - 1}\}$ of length $T_a$, and $\bm{o}^\tau$ is the corresponding observation at the begin of the target action series.\footnote{In this paper, the subscript $t$ represents the time step of the FM generation process, and the superscript $\tau$ indicates the real-world time step of the robot trajectory.} 

The vector field network $v_\theta(\bm{A}, t; \bm{o})$ is learned using the following loss function,
\begin{equation}
    \label{equation: fmp loss function}
    \mathcal{L}_{\text{FMP}}(\bm{\bm{\theta}}) = \mathrm{E}_{t, q(\bm{A}_1), p(\bm{A}_0)}||v_\theta(\bm{A}, t; \bm{o}) - u_t(\bm{A}|\bm{z}, \bm{o})||^2 ,
\end{equation}
where $q(\bm{A}_1)$ and $p(\bm{A}_0)$ are the target and prior distributions, respectively.
The training process of FMP, summarized in Algorithm~\ref{algorithm: training}, consists of three steps: (1) Sample from the prior and target distributions; (2) Compute the target conditioned vector field using~\eqref{equation:fmp vector field}; (3) Update the neural network parameters $\bm{\theta}$ using the loss function~\eqref{equation: fmp loss function}.

After training the observation-conditioned vector field, the policy inference process boils down to integrating the learned vector field, starting from samples drawn from the prior Gaussian distribution. As the analytical solution to this integration is generally intractable, numerical methods are commonly employed. Among these, the Euler method~\cite{euler} provides a straightforward and effective approximation.
The Euler method approximates the integration of the vector field by discretizing the integration into small time steps. Starting with an initial sample $\bm{A}_0$ from the prior Gaussian distribution and observation $\bm{o}$ of the environment, the integration process iteratively updates the state as follows,
\begin{equation}
    \bm{A}_{t + \Delta t} = \bm{A}_t + v_\theta(\bm{A}, t;\bm{o})\Delta t,
\end{equation}
where $\Delta t$ is the integration step size and related to total number of integration steps $N$ via $\Delta t= 1 / N$. 
The final $\bm{A}_1$, generated at $t=1$ is the action series. Finally, each action is given sequentially to the robot to perform the task.

\subsection{Safe Policy Generation with PF2MP}
\label{subsec: pf}
The base FMP primarily focuses on replicating expert demonstrations to reproduce observed behaviors. However, it overlooks critical considerations, such as the safety of the generated action sequences, which are essential for real-world deployment, especially in obstacle-rich environments.

Expert demonstrations inherently encode more than just the execution of complex policies to accomplish specific tasks; they also implicitly capture valuable information about the environment's constraints (e.g., obstacles). The regions explored by the expert policies represent safe areas, while areas not traversed by the demonstrations often correspond to unsafe or high-risk regions. By neglecting this implicit safety information, previous algorithms risk generating policies that, although behaviorally accurate, may lead to unsafe or infeasible actions when deployed in obstacle-rich environments.

\begin{algorithm}[tp]
\linespread{1.15}\selectfont 
\caption{Potential Field-Guided Flow Matching Inference}
\renewcommand{\algorithmicrequire}{\textbf{Input:}}
\renewcommand{\algorithmicensure}{\textbf{Output:}}
\label{algorithm: safe infer}
\begin{algorithmic}[1]
\REQUIRE Trained vector field $v_\theta(\bm{A}, t; \bm{o})$, number of integration steps $N$, observation $\bm{o}$, prior distribution $p_0$, hyperparameter $\lambda$, potential field $\Phi(\bm{A})$
\ENSURE Generated action series $\bm{A}_1$.  
\STATE Compute the timestep $\Delta t = \frac{1}{N}$;
\STATE Sample from prior distribution $\bm{A}_0 \sim p_0$;
\FOR{$i =1:N$}
    \STATE Integrate the vector field learned by FMP as \\
    $\bm{A}_{t_i} = \bm{A}_{t_{i-1}} + \left [v_\theta(\bm{A}_{t_{i-1}}, t_{i-1}; \bm{o}) + \lambda \Phi(\bm{A}_{t_{i-1}})\right ] \Delta t$ ~\eqref{eq:PF2MP_inference} 
\ENDFOR
\RETURN $\bm{A}_1= \left \{\bm{a}^\tau, \bm{a}^{\tau + 1},..., \bm{a}^{\tau + T_a - 1} \right \}$
\end{algorithmic}
\end{algorithm}

To address this limitation, we propose the Potential Field-Guided Flow Matching Policy (PF2MP), which leverages the inherent constraint information embedded in the demonstration data. Specifically, our approach estimates the density $\hat{p}(\bm{a})$ of the demonstration trajectories via KDE and constructs a potential function $\phi(a)$ based on the estimated density as,
\begin{equation}
    \phi(\bm{a}) = \log(\hat{p}(\bm{a})) + \alpha d(\bm{a}, \mathcal{H}),
\end{equation}
which consists of 2 components, whose contributions are balanced by the hyperparameter $\alpha$ with $\alpha < 0$. The first term, $\log(\hat{p}(\bm{a}))$, represents the logarithm of the estimated density, while the second term, $d(\bm{a}, \mathcal{H})$, is a distance-based penalty that measures the proximity of an action $\bm{a}$ to the safer area $\mathcal{H}$. The safer area is defined as the region where the estimated density exceeds a predefined threshold, indicating regions covered by the expert demonstrations.
\begin{figure*}[t]
  \centering
  \begin{subfigure}[t]{0.18\linewidth}
    \centering
    \includegraphics[width=\linewidth, height=\linewidth]{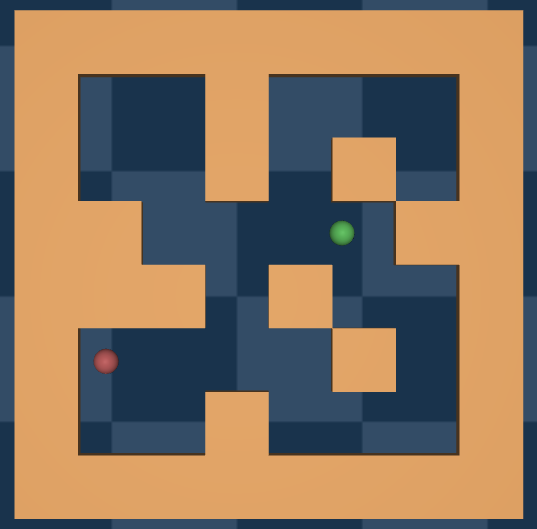}\\[0.5em]
    \includegraphics[width=\linewidth, height=\linewidth]{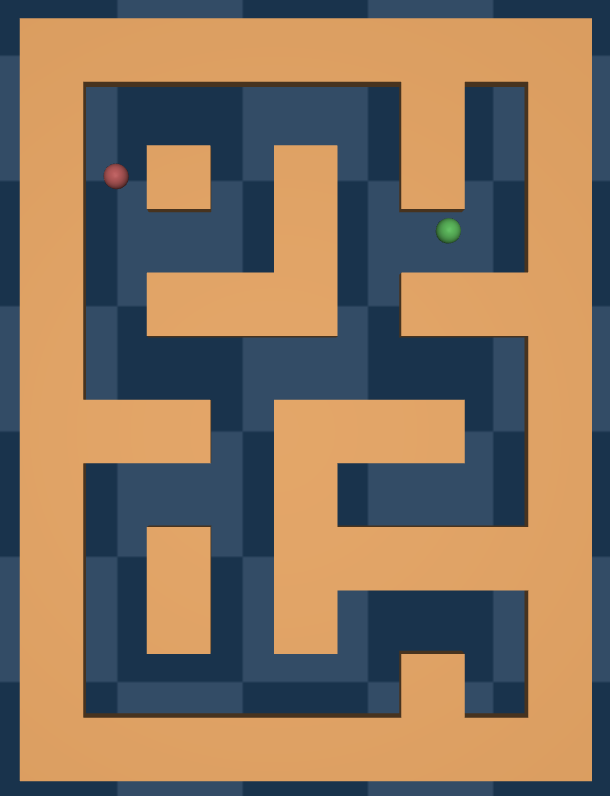}
    \caption{Gym Environment}
    \label{subfig: gym env}
  \end{subfigure}
  \hfill
  \begin{subfigure}[t]{0.18\linewidth}
    \centering
    \includegraphics[width=\linewidth, height=\linewidth]{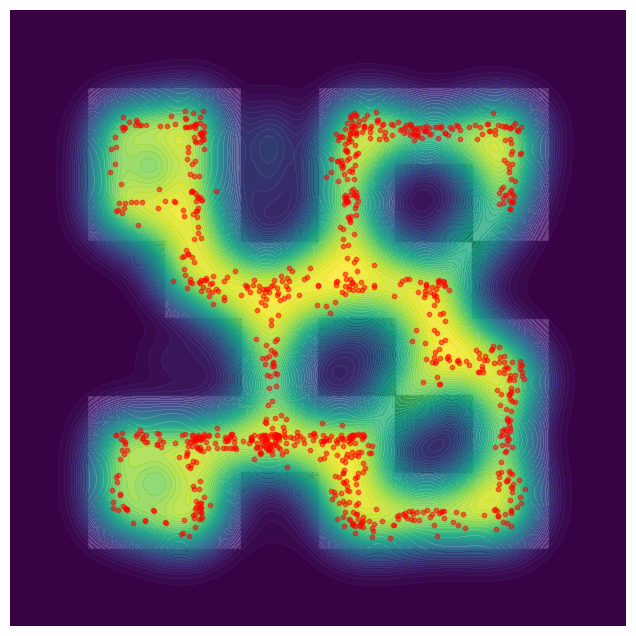}\\[0.5em]
    \includegraphics[width=\linewidth, height=\linewidth]{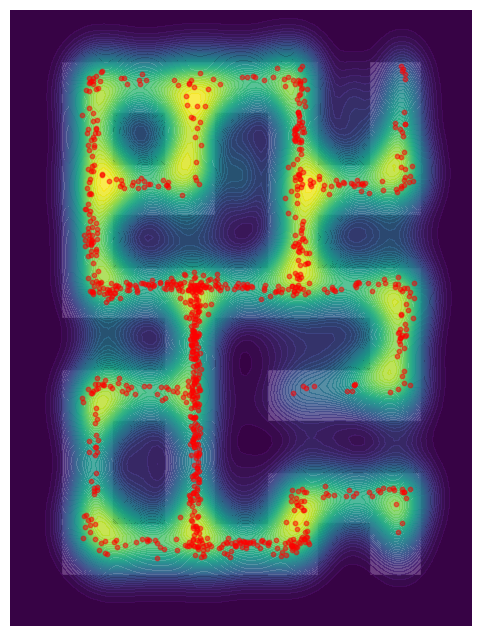}
    \caption{Density Map}
    \label{subfig: density map}
    
  \end{subfigure}
  \hfill
  \begin{subfigure}[t]{0.18\linewidth}
    \centering
    \includegraphics[width=\linewidth, height=\linewidth]{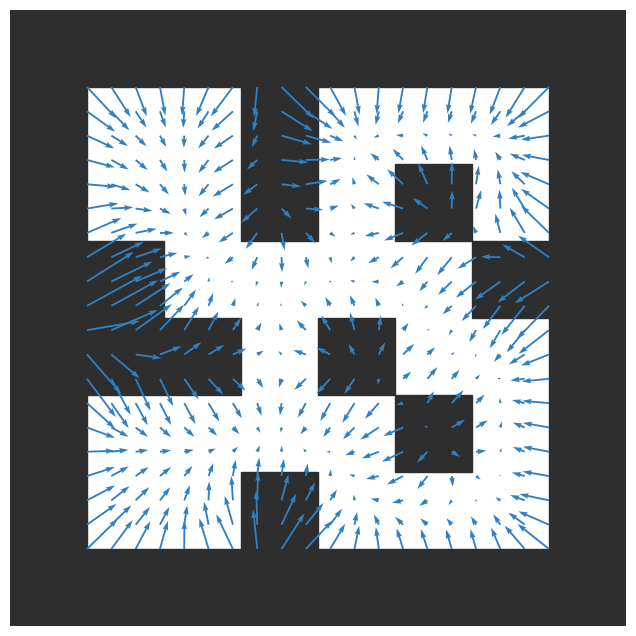}\\[0.5em]
    \includegraphics[width=\linewidth, height=\linewidth]{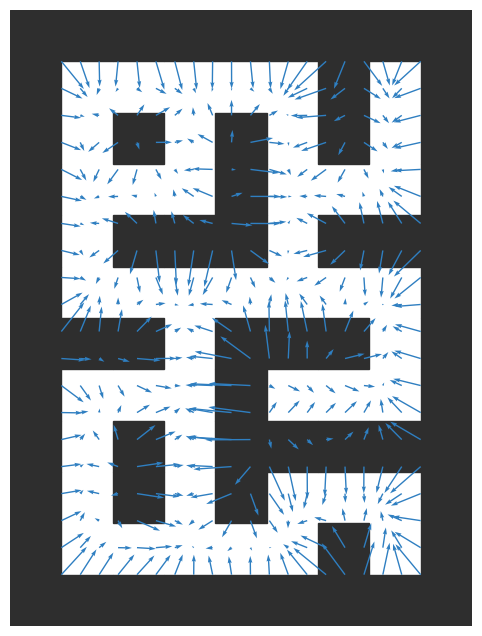}
    \caption{Potential Field}
    \label{subfig: potential field}
    
  \end{subfigure}
  \hfill
  \begin{subfigure}[t]{0.18\linewidth}
    \centering
    \includegraphics[width=\linewidth, height=\linewidth]{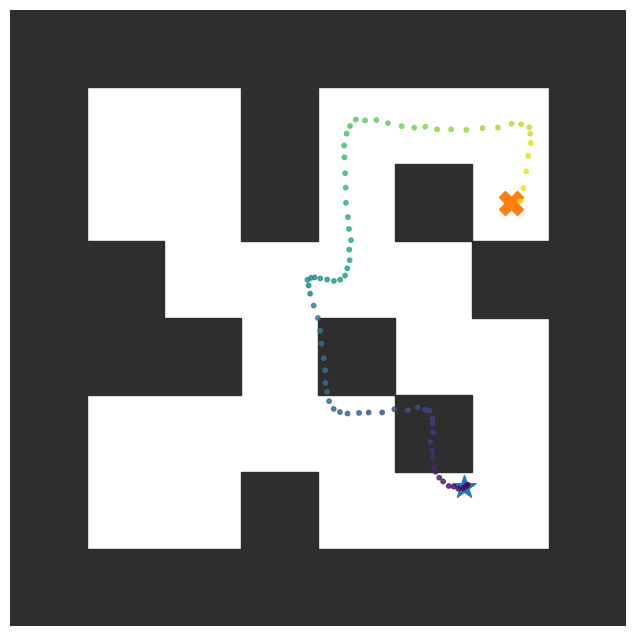}\\[0.5em]
    \includegraphics[width=\linewidth, height=\linewidth]{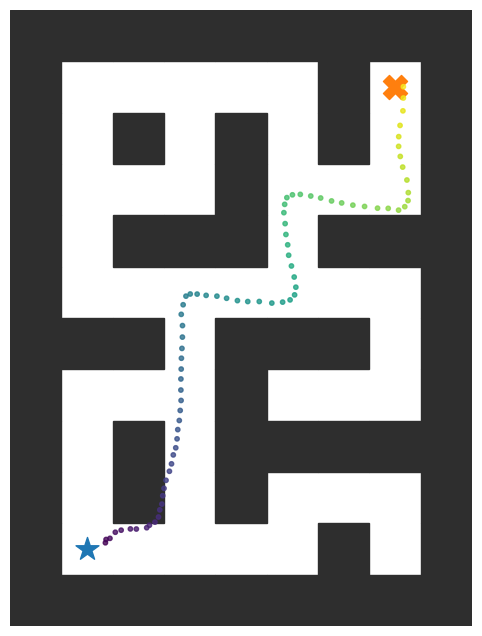}
    \caption{FMP}
    \label{subfig: fmp}
    
  \end{subfigure}
  \hfill
  \begin{subfigure}[t]{0.18\linewidth}
    \centering
    \includegraphics[width=\linewidth, height=\linewidth]{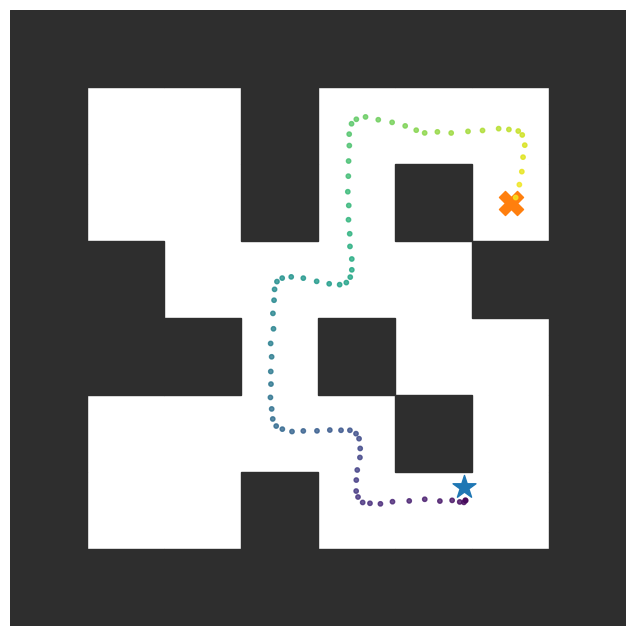}\\[0.5em]
    \includegraphics[width=\linewidth, height=\linewidth]{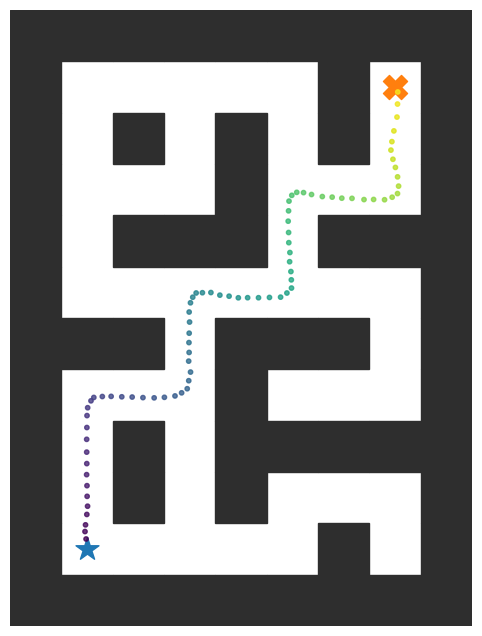}
    \caption{PF2MP}
    \label{subfig: pf2mp}
    
  \end{subfigure}
  \vspace{0.2cm}
  \caption{2D maze navigation tasks: (a) Gym environment with the medium maze (top row) and large maze (bottom row). (b) Estimated density map, with red points representing the demonstration data. (c) Derived potential field, visualized with blue arrows, based on the estimated density. (d) Failure case of the baseline FMP, where collisions occur, with the blue star indicating the start point and the orange fork marking the goal. (e) Collision-free trajectory generated by the PF2MP under the same initial and target conditions as in (d).}
    \label{fig:2D maze}
\end{figure*}

The resulting potential field $\Phi(A)$ is the gradient of the potential function across the action series, i.e.,
\begin{equation}
    \Phi(\bm{A}) =  \left[\frac{d\phi}{d\bm{a}^\tau},  \frac{d\phi}{d\bm{a}^{\tau + 1}}, ..., \frac{d\phi}{d\bm{a}^{\tau + T_a-1}}\right].
    \label{eq:PF2MP_inference}
\end{equation}

The policy inference process of PF2MP integrates the learned flow matching vector field with the designed potential field into a unified framework. As summarized in Algorithm~\ref{algorithm: safe infer}, the procedure begins by sampling an initial point $\bm{A}_0$ from a prior Gaussian distribution, representing the initial state of the generated action series. At each subsequent integration step, the intermediate state $\bm{A}_t$ is updated by following the learned FM vector field $v_\theta(\bm{A}, t|\bm{o})$, which has been trained to mimic the expert demonstrations, augmented with the potential field $\Phi(\bm{a})$, constructed from the density estimation of the demonstration data. The potential field acts as a corrective term that incorporates safety constraints by steering the generated actions away from low-density (unsafe) regions and toward high-density (safe) regions. Mathematically, the update at each integration step is expressed as,
\begin{equation}
    \bm{A}_{t + \Delta t} = \bm{A}_{t} + [v_\theta(\bm{A}_{t}, t; \bm{o}) + \lambda \Phi(\bm{A}_{t})] \Delta t ,
\end{equation}
where $\lambda$ is a hyperparameter controlling the trade-off between flow matching dynamics and potential field guidance, and $\Delta t$ represents the integration step size of the Euler method. 
This iterative process is repeated $N$ times, where $N$ is a hyperparameter determining the resolution of the integration. By integrating safety-aware potential fields into the flow matching framework, PF2MP generates action sequences that not only replicate expert behavior but also account for the obstacles present in the demonstrations, ensuring both effectiveness and safety in complex environments.

\begin{table*}[t]
    \centering
    \caption{Task performance of different models on 2D maze navigation and Fetch robot manipulation tasks.}
    \label{table: maze & fetch}
    \begin{tabular}{lcccc|cccc}
        \toprule
        \multirow{2}{*}{Tasks} & \multicolumn{4}{c}{Success Rate (\%)} & \multicolumn{4}{c}{Collision Rate (\%)} \\
        \cmidrule(lr){2-5} \cmidrule(lr){6-9}
                              & FMP   & \textbf{PF2MP}   & DP   & DPPF  & FMP   & \textbf{PF2MP}   & DP   & DPPF  \\
        \midrule
        Maze Medium               &  $\bm{99.7 \pm 0.1}$  &  $99.6 \pm 0.2$  &  $99.6 \pm 0.2$  &  $99.6 \pm 0.2$  &
                              $ 0.33 \pm 0.12$& $ \bm{0.07 \pm 0.07}$  &  $0.47 \pm 0.12$ &  $0.27 \pm 0.12$ \\
        Maze Large           & $96.6 \pm 1.1$    & $96.4 \pm 1.2$   &  $\bm{96.7 \pm 0.8}$   &  $96.4 \pm 1.2$      & $9.2 \pm 0.8$  & $ \bm{3.2 \pm 0.5}$  & $8.8 \pm 0.7 $  &   $7.6 \pm 0.5$  \\
        Fetch Reach              &  $\bm{85.0 \pm 4.4}$  &  $84.0 \pm 1.7$  &  $84.0 \pm 3.5$   &  $83.0 \pm 2.0$  &  $34.0 \pm 5.6$  &  $\bm{18.0 \pm 4.4}$  &  $35.0 \pm 4.4$ & $33.0 \pm 3.0$ \\
        Fetch Push                &  $50.0 \pm 4.6$   &  $\bm{51.7 \pm 2.5}$  &  $48.3 \pm 2.5$  &  $47.0 \pm 3.6$  &  
                        $30.0 \pm 3.6$&  $\bm{25.0 \pm 2.6}$   &  $33.3 \pm 5.7$  &  $30.3 \pm 4.7$  \\
        \bottomrule
    \end{tabular}
\end{table*}

\section{EXPERIMENT}
We conducted four sets of experiments, including navigation and robot manipulation tasks in simulated environments, as well as a real-world robotic task. These experiments are designed to comprehensively evaluate the effectiveness of the proposed PF2FMP in enhancing safety with task execution controlled in task space or joint space.

\subsection{Basic Settings}
We implement a conditional 1D U-Net similar to~\cite{diffusionpolicy, haoran2024srfmp}, as the base network for learning the vector field of the base FM policy. The U-Net consists of a three-layer structure with hidden dimensions [512, 256, 128], designed to efficiently model the observation-conditioned vector field. For comparison, we evaluate PF2MP against three baselines: base Flow Matching Policy (FMP), Diffusion Policy (DP), and DP with Potential Field Guidance (DPPF) as a modified baseline, where the same potential field guidance employed in PF2MP is integrated into the denoising process of DP.

To measure the performance of different policies, we consider the collision rate, given as the percentage of trajectories colliding with obstacles in the environment. For the 2D Maze (Section~\ref{subsec:2Dmaze}) and Fetch (Section~\ref{subsec:fetch}) tasks, we also consider the task success rate, irrespectively of collisions during the execution.

\subsection{2D Maze Navigation Tasks}
\label{subsec:2Dmaze}
We begin by evaluating our approach on a navigation task in the 2D Maze environment provided by D4RL~\cite{d4rl, anonymous2024gymnasium}, which is built upon the Mujoco physics engine~\cite{todorov2012mujoco}. As illustrated in Fig~\ref{fig:2D maze}, the environment consists of brown obstacles and boundaries. The controllable agent is represented by a green ball, whose objective is to navigate through the maze to reach a red ball. The training dataset, provided by D4RL, includes trajectories with random start and goal states within the safe region. To ensure trajectory smoothness, we resample the demonstrations to a fixed length of $80$ steps, which is sufficient to complete the task effectively. D4RL provides mazes with varying levels of complexity. For our experiments, we select the ``medium'' and ``large'' maze configurations.
In this task, we utilize PF2MP to predict the trajectory of the green ball, assuming the existence of a low-level controller capable of precisely reaching the predicted states. The observation $\bm{o}$ is the position of the green and red balls. To construct the potential field, we randomly sample $3,000$ points from the demonstrations instead of using the entire dataset to reduce computational overhead. 

We evaluate PF2MP and the baseline methods using the same $500$ pairs of initial and target states, repeating each test three times. 
All models are inferred with $N=5$ integration steps. The weight $\lambda$ for PF2MP and DPPF is set to $0.8$. 
\begin{figure}[t]
    \centering
    \includegraphics[width=0.7\linewidth]{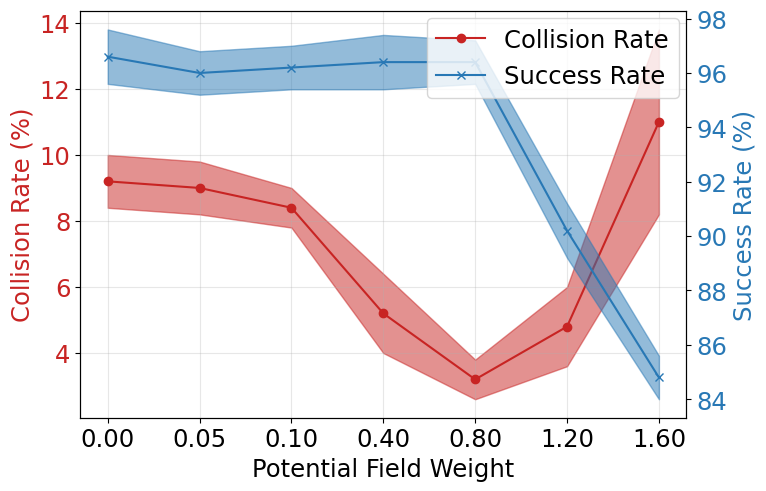} 
    \caption{Impact of the potential field weight $\lambda$ on PF2MP performance in the large maze task. 
    }
    \label{fig:ablation potential field weight}
    \raggedbottom
\end{figure}

The performance of all models are reported in Table~\ref{table: maze & fetch}.
In the medium maze, all models exhibit strong performance, with collision rates below $0.4\%$ and success rates exceeding $99\%$. PF2MP achieves a marginally lower collision rate compared to FMP while maintaining an equivalent success rate. A similar trend is observed with DPPF compared to DP, where a slight reduction in collision rate is evident.
In the large maze, PF2MP demonstrates a significant reduction in collision rate compared to FMP, achieving the lowest collision rate ($\sim3\%$) among all models, while maintaining a comparable success rate to other baselines. These results indicate that the potential field guidance enhances the safety of the generated trajectories without adversely affecting task performance. This improvement is consistently observed across both FMP and DP, underscoring the general effectiveness of the potential field guidance.

To further analyze the effect of the potential field guidance, we conducted an ablation study on the potential field weight $\lambda$ using the large maze setting (see results in Fig~\ref{fig:ablation potential field weight}). When $\lambda$ is set too low ($\lambda<0.1$), the performance of PF2MP is nearly identical to the baseline FMP. As $\lambda$ increases to values such as $0.4$ and $0.8$, the collision rate significantly decreases, and the task success rate remains stable. However, when $\lambda$ is further increased, the collision rate rises dramatically, and the success rate decreases accordingly. We hypothesize that this occurs because a high $\lambda$ leads to too-narrow safe area: As the agent tries to follow the demonstrated trajectory, it overshoots and collides with the opposite obstacle. This phenomenon suggests that while PF2MP can enhance safety in action generation, careful selection of the $\lambda$ parameter is crucial for optimal performance. 
\subsection{Simulated Robotics Tasks in Task Space}
\label{subsec:fetch}
\begin{figure}[t]
    \centering
    \includegraphics[width=0.45\linewidth]{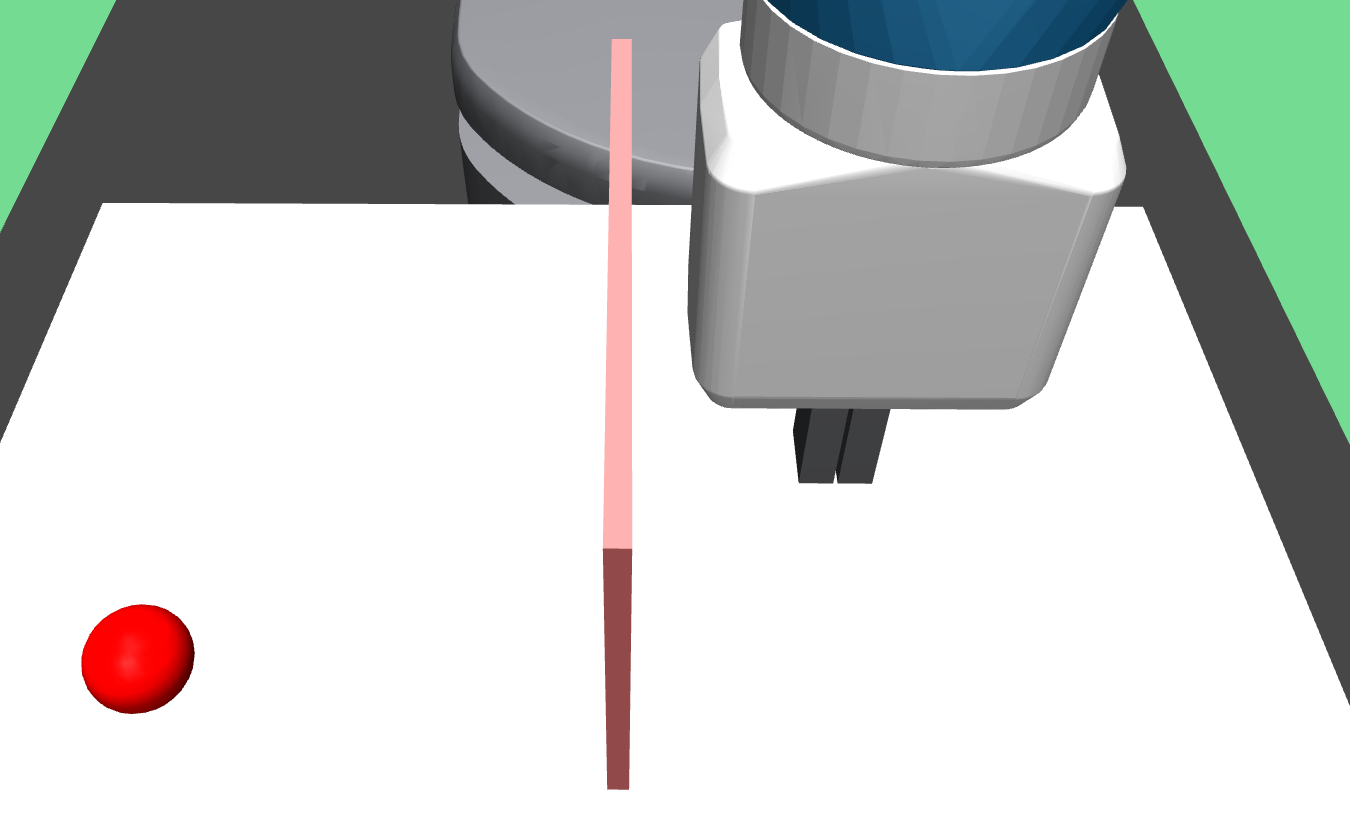} 
    \includegraphics[width=0.45\linewidth]{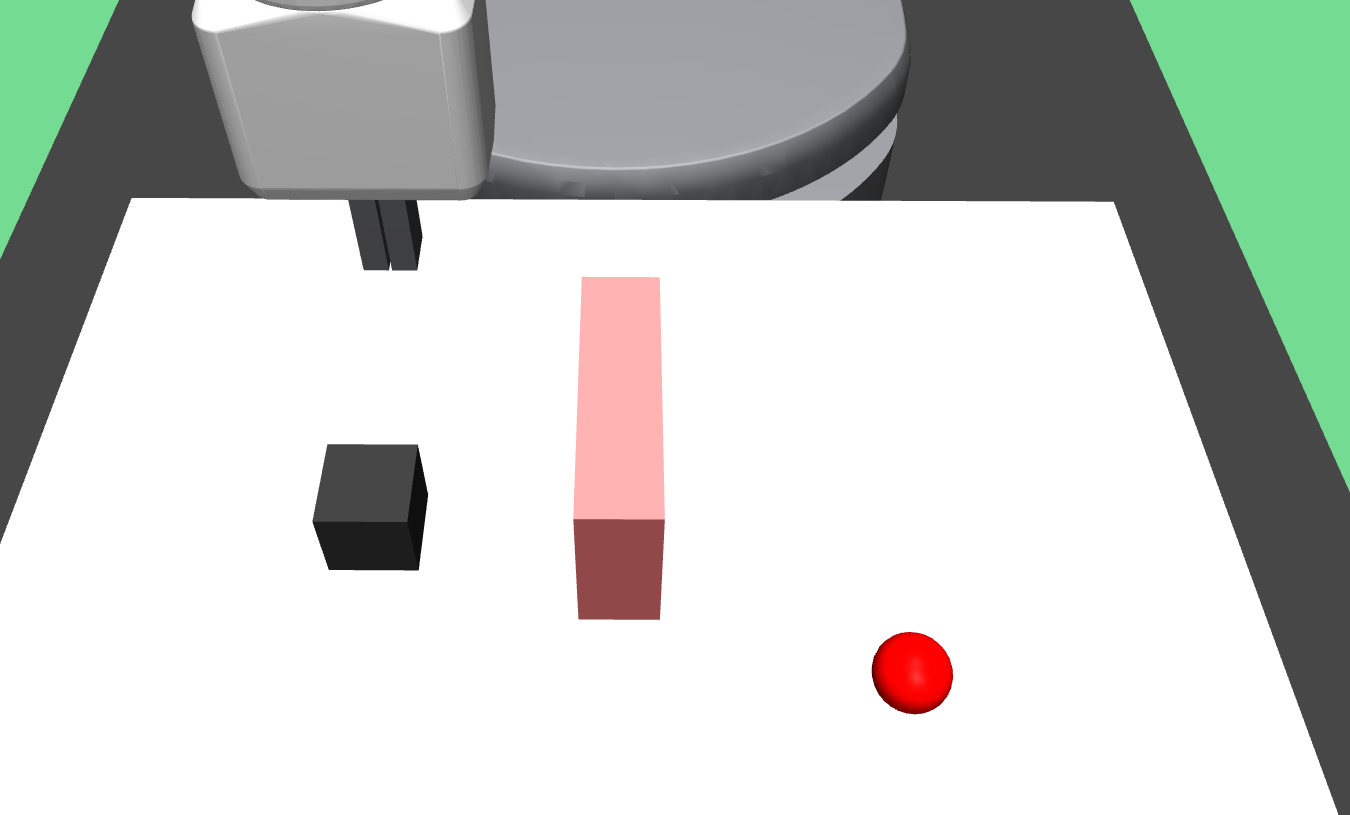} 
    \caption{Fetch Environments. Left: Reach task. 
    Right: Push task. 
    }
    \label{fig:fetch env}
\end{figure}

We evaluate PF2MP on robot manipulation tasks by considering two benchmark tasks --- Reach and Push --- performed using the Fetch robot simulator from Gymnasium Robotics~\cite{fetchsim}. To evaluate the safety of the proposed method, we augment the original Fetch environment by introducing obstacles, as shown in Fig~\ref{fig:fetch env}. This setup is inspired by and adapted from the experimental framework presented in~\cite{colddiffusion}. 
In the Reach task, a thick red wall is placed as an obstacle in the middle of the platform. The robot’s end-effector is initialized on the right side of the wall, while the target goal, represented by a red ball, is randomly initialized on the left side of the wall. The objective is to control the robot to navigate around the obstacle and reach the goal point.
In the Push task, a similar thick red wall is present in the middle of the platform, but with a reduced height compared to the one used in the Reach task. The objective is to control the robot to push a black block to a target position marked by a red ball. Both the black block and the red ball are initialized on the surface of the platform.

To collect expert demonstrations, we employ the Stable-Baselines3, a Python library~\cite{stable-baselines3} for Reinforcement Learning. For the Reach task, we train a reinforcement learning agent using Proximal Policy Optimization (PPO)~\cite{ppo}. For the Push task, we employ Soft Actor-Critic (SAC)~\cite{sac} combined with Hindsight Experience Replay (HER)~\cite{her} as the replay buffer. Both tasks are trained for a total of $1 \times 10^6$ time steps. 
After training, the RL agents are used to collect $1000$ expert demonstrations for each task under varying initial conditions. To standardize the dataset, the length of each demonstration is resampled to $48$ steps, which is sufficient to successfully complete both tasks.

The performance of our proposed method, PF2MP, along with other baseline models, is summarized in Table~\ref{table: maze & fetch}. For the Reach task, all models achieve an average success rate of approximately $84\%$. However, PF2MP stands out as the only method with a collision rate below $20\%$, specifically $18\%$, which is nearly half the collision rate of the baseline FMP at $34\%$. For the DP model, the potential field guidance does not demonstrate significant effectiveness compared to the 2D Maze scenario, with DPPF outperforming DP by only $2\%$.
For the Push task, which is simplified as a 2-points reach task, the overall performance of all models is less robust, achieving an average success rate of only around $50\%$, with collision rates hovering at approximately $30\%$. PF2MP shows a modest improvement in collision rate, reducing it by approximately $5\%$. This may be attributed to the low quality of the demonstration data collected by the RL agent, which tends to favor trajectories that closely follow obstacles.  Additionally, the potential field guidance contributes a slight increase in success rate for this task, achieving a gain of about $1.7\%$.

\begin{figure}[t]
    \centering
    \includegraphics[width=0.7\linewidth]{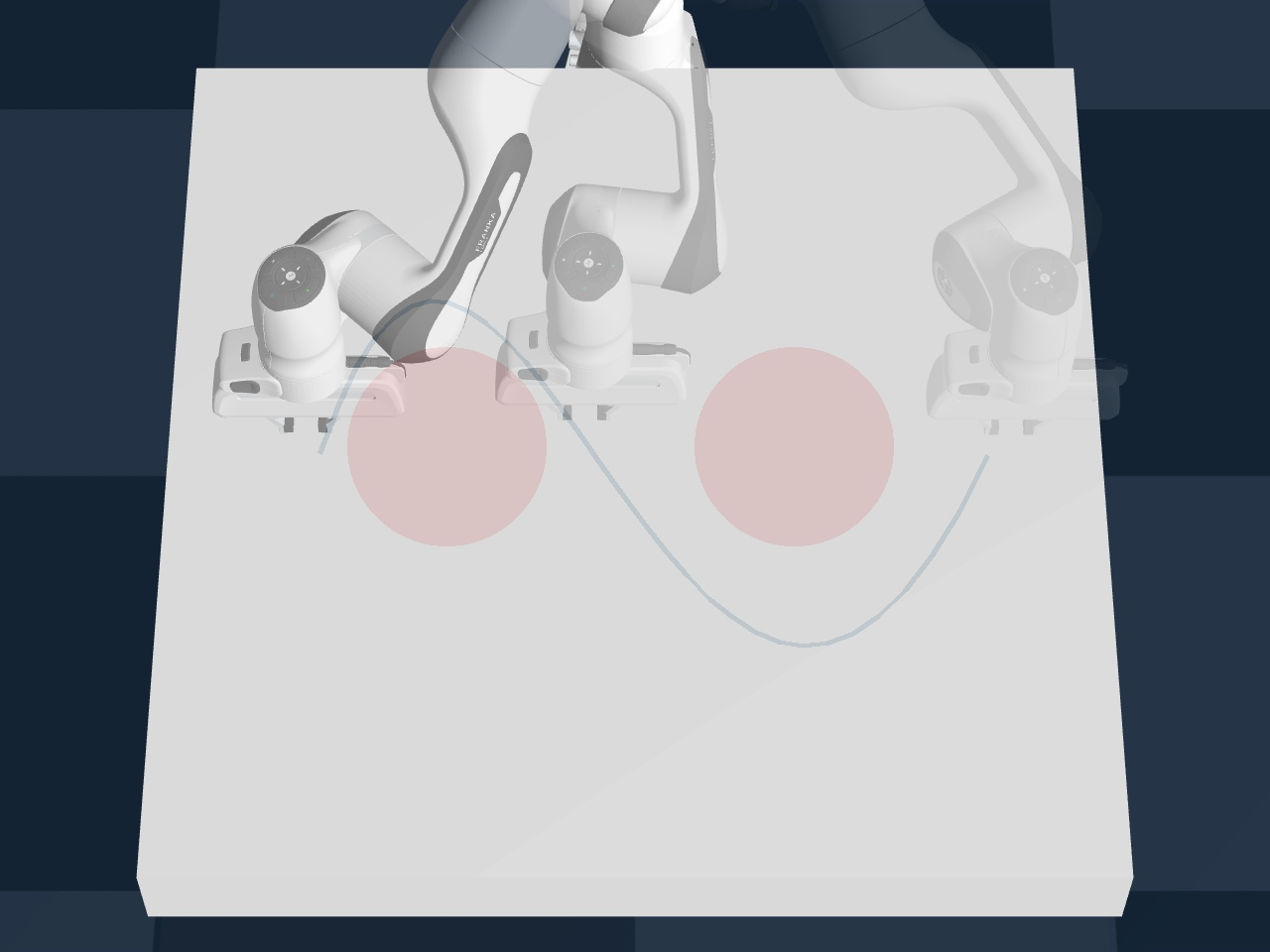}  
    \includegraphics[width=0.25\linewidth, height=0.5\linewidth, keepaspectratio]{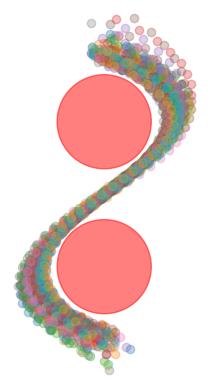} 
    \caption{Genesis Handwriting Task. Left: Simulation environment in Genesis, featuring a Franka Panda robotic arm, a white tabletop, and an ``S''-shaped trajectory that avoids the red obstacle regions. Right: 2D end-effector trajectories obtained from the joint-space demonstrations.}
    \label{fig:genesis}
\end{figure}

\subsection{Simulated Robotics Tasks in Joint Space}
To validate the effectiveness of our proposed approach to learn policies in joint space, we designed a handwriting experiment using the Genesis simulator~\cite{genesis}. In this experiment, a Franka Panda robotic arm was controlled to draw letters on a planar surface while avoiding predefined restricted regions, as illustrated in Fig~\ref{fig:genesis}. Unlike our previous experiments, this task involves direct control of the robot's joint space, encompassing $9$ degrees of freedom ($7$ for the revolute joints, $2$ for the gripper). The experimental setup highlights the adaptability of our method in generating safe and feasible trajectories under constraints, even in complex and high-dimensional task spaces.
\begin{table}[t]
\centering
\vspace{0.2cm}
\caption{Handwriting Tasks Collision Rate}
\begin{tabular}{lcc}
\toprule
\textbf{Policies} & \multicolumn{2}{c}{\textbf{Collision Rate}} \\
 & Genesis Simulator & Unitree Z1 Real Robot\\
\midrule
FMP  &$25.33 \pm 2.52 (\%)$ & $4/10$ \\
PF2MP & $0.67 \pm 0.58(\%)$ & $0/10$\\
\bottomrule
\end{tabular}
\label{table:hand write}
\end{table}

To train the policies, we collected $20$ demonstrations with the corresponding end-effector trajectories illustrated in Fig~\ref{fig:genesis}. The observation $\bm{o}$ in this task is the state of the robot arm in joint space. During testing, all policies were also inferred with $N=5$ integration steps. The PF2MP weight parameter $\lambda$ was set to $0.5$. Every policy was tested using the same set of $100$ prior samples, and this evaluation was repeated three times for statistical robustness. The results, summarized in Table~\ref{table:hand write}, demonstrate a clear advantage of our proposed PF2MP method in generating safer trajectories. Specifically, the baseline FMP approach exhibited a collision rate of approximately $25\%$, whereas our method achieved a substantial reduction in the collision rate to just $0.67\%$. This significant improvement demonstrates the ability of PF2MP to directly predict safe trajectories in the joint space.

\subsection{Real Robotics Tasks}
\begin{figure}[t]
    \centering
    \includegraphics[width=0.23\linewidth, height=0.45\linewidth]{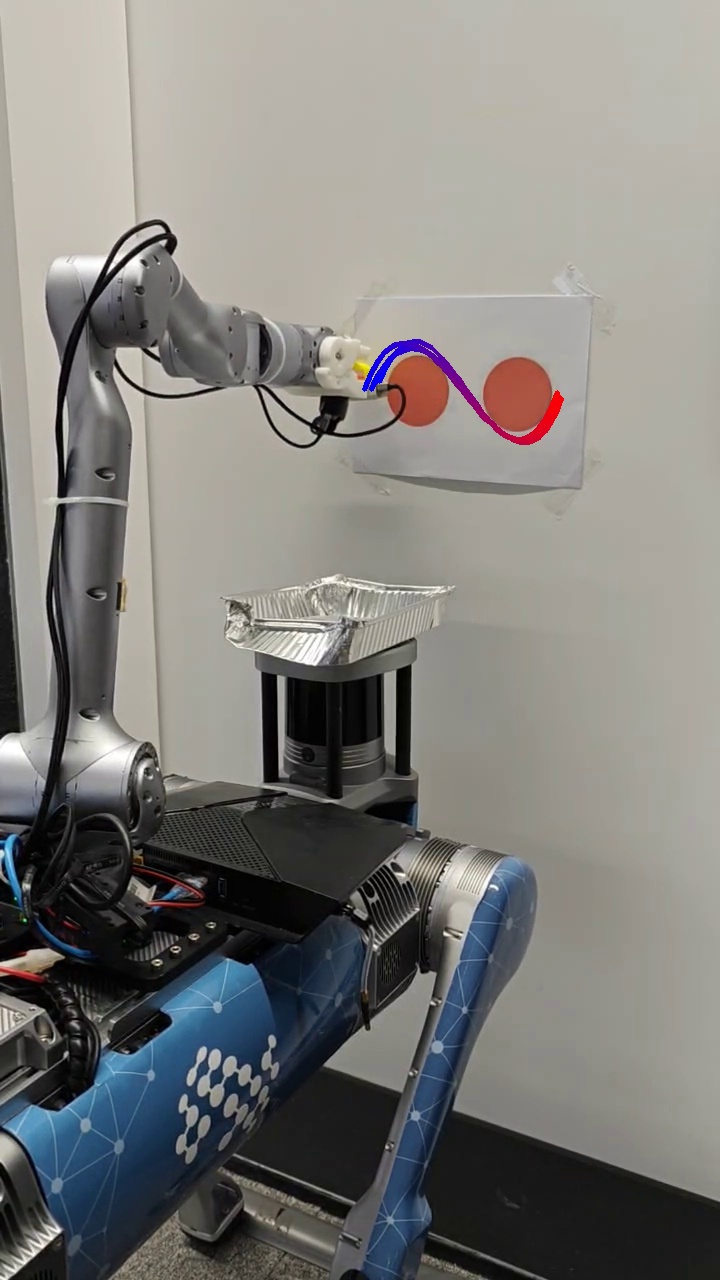} 
    \includegraphics[width=0.23\linewidth, height=0.45\linewidth]{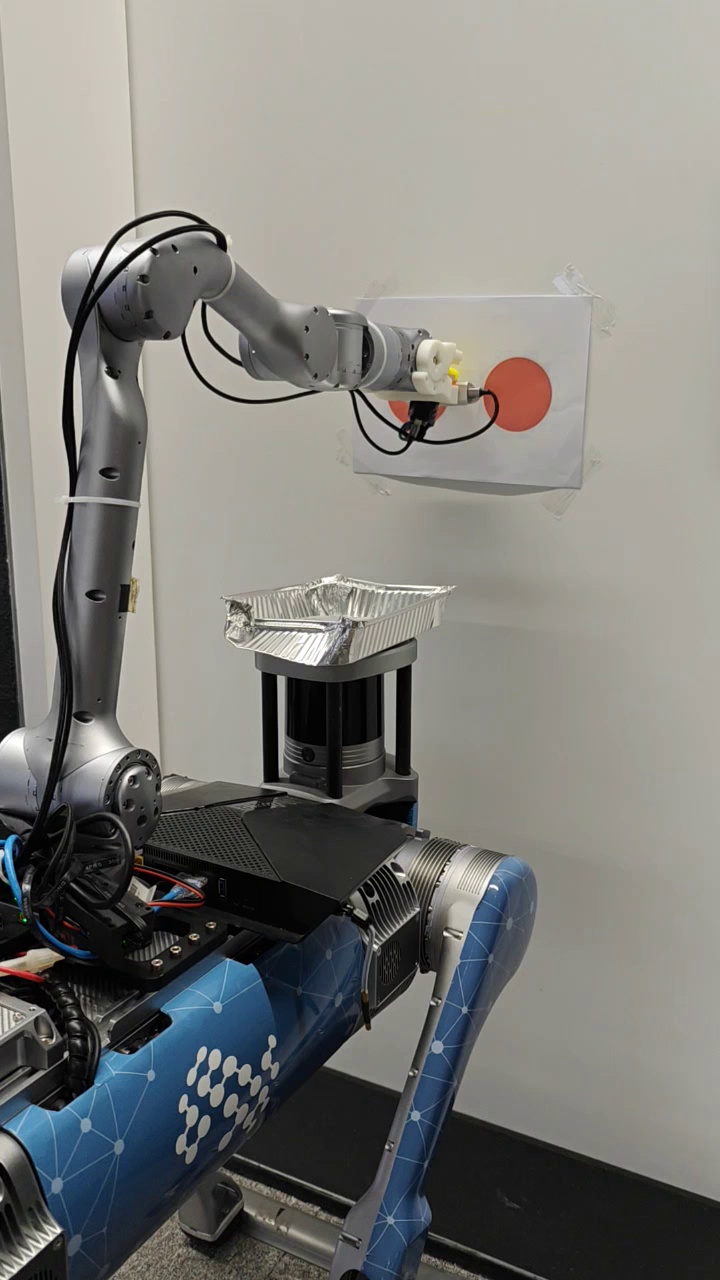} 
    \includegraphics[width=0.23\linewidth, height=0.45\linewidth]{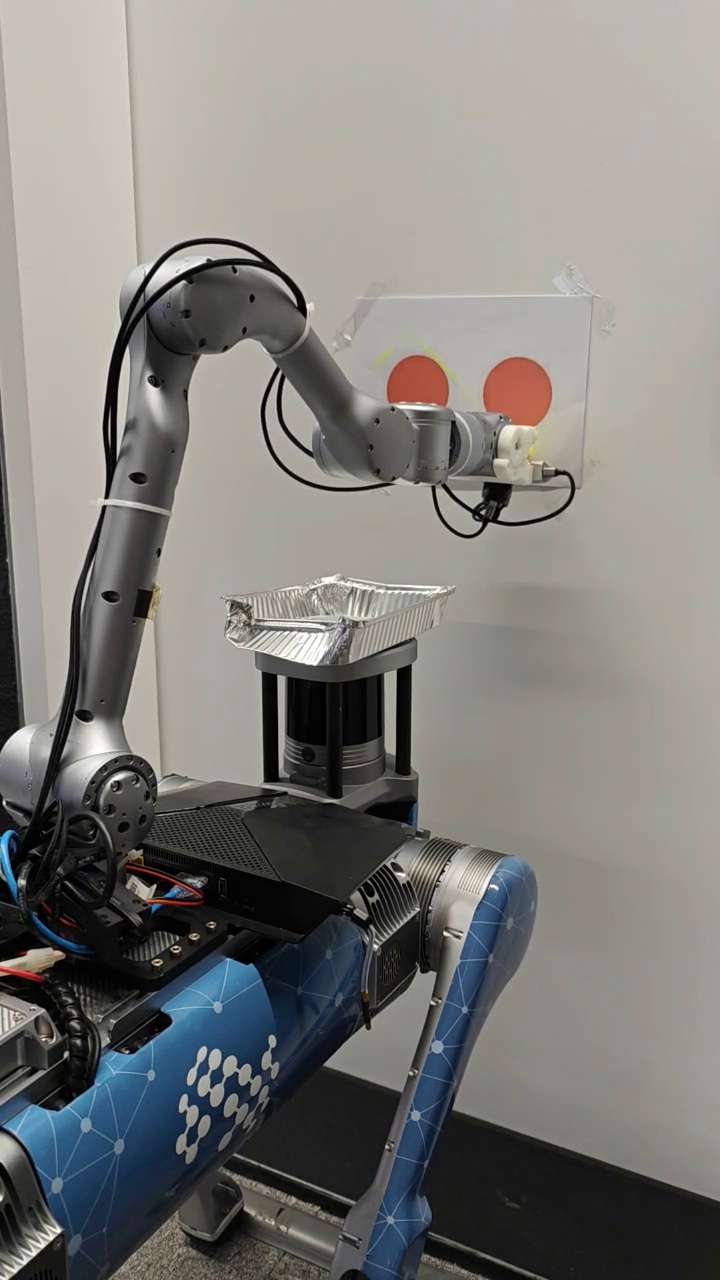} 
     \includegraphics[width=0.23\linewidth, height=0.45\linewidth]{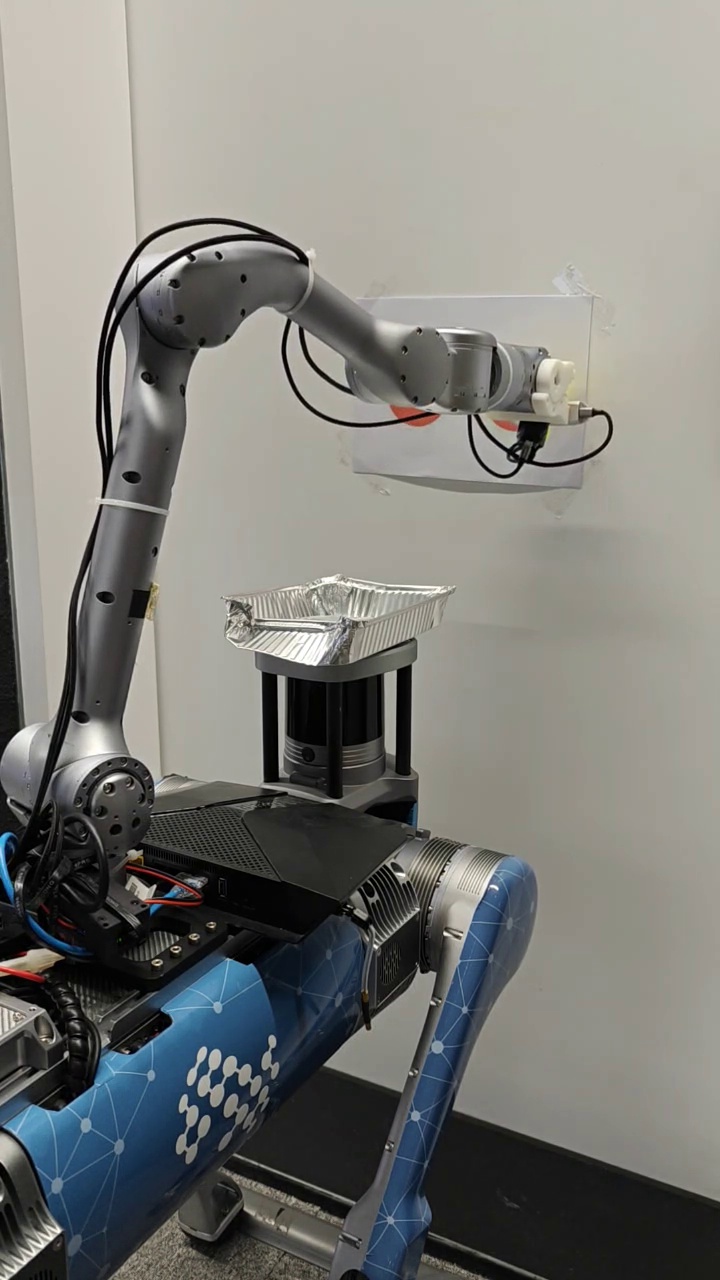} 
    \caption{Real-World Handwriting Task. A Unitree Z1 robotic arm mounted on a Unitree B2 robot dog draws the letter ``S'' on a wall-mounted sheet of paper, while ensuring that the trajectory avoids the predefined pink constrained area on the paper.}
    \label{fig:real robot}
\end{figure}

Finally, we evaluate PF2MP in a real-world scenario for a task executed on the Unitree Z1 robot arm mounted on a Unitree B2 base, as shown in Fig~\ref{fig:real robot}. This experiment replicates the Genesis Handwriting task but planning in task space, where the robot arm is tasked with plotting the letter ``S'' on a planar surface while avoiding a predefined constraint region, indicated by the pink circle in Fig~\ref{fig:real robot}. For this experiment, we collected $10$ demonstrations with varying start and end points of the trajectory, which were generated via teleoperation using a keyboard interface provided by Unitree's official software package. We employed a smaller UNet with down dimension $[16, 32, 64]$, training it for $100$ epochs. The observation $\bm{o}$ in this task is the state of robot end effector, and the prediction is the trajectory of the robot end effector.

During testing, the hyperparameter $\lambda$ for PF2MP was set to $0.5$. Both policies were tested $10$ times under identical initial conditions. The results, presented in Table~\ref{table:hand write}, clearly demonstrate the safety advantages of PF2MP. For the FMP baseline, approximately $4$ out of the $10$ trials resulted in trajectories that crossed the pink constraint area. In contrast, trajectories generated by PF2MP consistently avoided the constraint area in all trials, demonstrating its superior ability to generate safe trajectories under real-world conditions. 

\section{CONCLUSIONS}
In this work, we introduced Potential Field-Guided Flow Matching Policy (PF2MP), a novel approach that jointly learns a complex policy and an associated potential field from the same set of demonstrations. By extending the capabilities of existing FM-based imitation learning methods, PF2MP generates safer policies that significantly reduce collision rates. Extensive evaluations across diverse simulated and real-world robotics tasks demonstrated that our approach leads to a substantial reduction in collisions.

Despite its promising performance, PF2MP exhibits sensitivity to potential field weight, which requires careful tuning to balance safety and task success. Additionally, our experiments are currently restricted to static environments with stationary obstacles. Addressing these limitations, future work could focus on designing a framework that incorporates dynamic obstacle handling by learning a dynamic potential field, paving the way for broader applicability and improved robustness in real-world scenarios.






\bibliographystyle{IEEEtran}
\balance
\bibliography{literature}

\end{document}